\documentclass[conference]{IEEEtran}
\usepackage{graphicx}
\usepackage{subfig}
\usepackage{graphicx}
\usepackage{amsmath}
\usepackage{multirow}
\usepackage{color,xcolor}
\usepackage{lipsum}

\ifCLASSINFOpdf
\else
\fi
\begin{document}
\title{Scene Text Magnifier}


%\author{\IEEEauthorblockN{Anna Zhu}
%\IEEEauthorblockA{School of Computer\\
%Wuhan University of Technology\\
%Wuhan, China\\
%Email: annakkk@live.com}
%\and
%\IEEEauthorblockN{Seiichi Uchida}
%\IEEEauthorblockA{Human Interface Laboratory\\
%Kyushu University\\
%Fukuoka, Japan\\
%Email: uchida@ait.kyushu-u.ac.jp}}
\author{\IEEEauthorblockN{Toshiki Nakamura\IEEEauthorrefmark{2},
Anna Zhu\IEEEauthorrefmark{1},and
Seiichi Uchida\IEEEauthorrefmark{2}}
\IEEEauthorblockA{\IEEEauthorrefmark{2}Human Interface Laboratory, Kyushu University, Fukuoka, Japan. Email: \{nakamura,uchida\}@human.ait.kyushu-u.ac.jp}
\IEEEauthorblockA{\IEEEauthorrefmark{1}School of Computer, Wuhan University of Technology, Wuhan, China. Email: annakkk@live.com(Corresponding Author)}}

\maketitle
%\newcommand\blfootnote[1]{%
%  \begingroup
%  \renewcommand\thefootnote{}\footnote{#1}%
%  \addtocounter{footnote}{-1}%
%  \endgroup
%}
%\blfootnote{*Anna Zhu is the Corresponding Author.} %

\begin{abstract}
Scene text magnifier aims to magnify text in natural scene images without recognition. It could help the special groups, who have myopia or dyslexia to better understand the scene. In this paper, we design the scene text magnifier through interacted four CNN-based networks: character erasing, character extraction, character magnify, and image synthesis. The architecture of the networks are extended based on the hourglass encoder-decoders. It inputs the original scene text image and outputs the text magnified image while keeps the background unchange. Intermediately, we can get the side-output results of text erasing and text extraction. The four sub-networks are first trained independently and fine-tuned in end-to-end mode. The training samples for each stage are processed through a flow with original image and text annotation in ICDAR2013 and Flickr dataset as input, and corresponding text erased image, magnified text annotation, and text magnified scene image as output. To evaluate the performance of text magnifier, the Structural Similarity is used to measure the regional changes in each character region. The experimental results demonstrate our method can magnify scene text effectively without effecting the background.

\end{abstract}


% \fi
\IEEEpeerreviewmaketitle

\section{Introduction}

Text, as a essential visual element, appears broadly in our daily life. They provide enrich information in different scenarios, such as the signs for guidance, the advertisements for goods and prices, the book title for searching, etc. Text in image can provide extra information of the scene, and assist the understanding of it, which can be used in a wide range of applied Computer Vision (CV) tasks. Typically, text are extracted from images through detection and recognition. On the other hand, it is desirable to magnify the text in the scene instead of recognition. For instance, those text which are far from the capturing position are relatively small resulting the difficulty for reading. Small characters printed on newspapers and other files are unobserving for group of myopic. It is more difficult for people who are effected by dyslexia. Therefore, magnifying the text in the image can assist better understanding of the information and make it possible to prevent missed reading and misunderstanding.

\begin{figure}
\centering
\subfloat[Detected text.]
{\label{fig1_1}
\includegraphics[width=3.2cm]{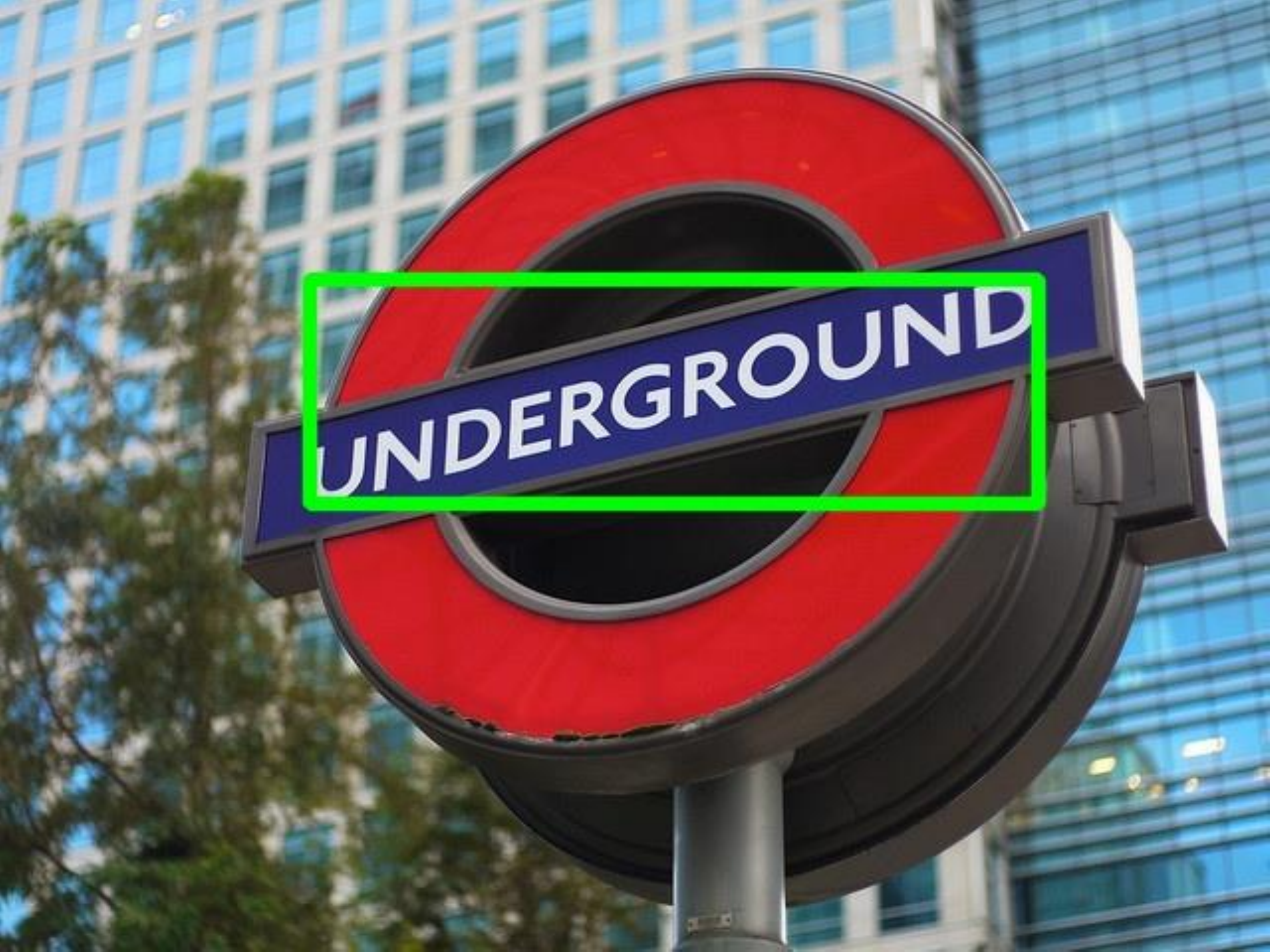}
}
\subfloat[Magnified text.]
{\label{fig1_2}
\includegraphics[width=3.2cm]{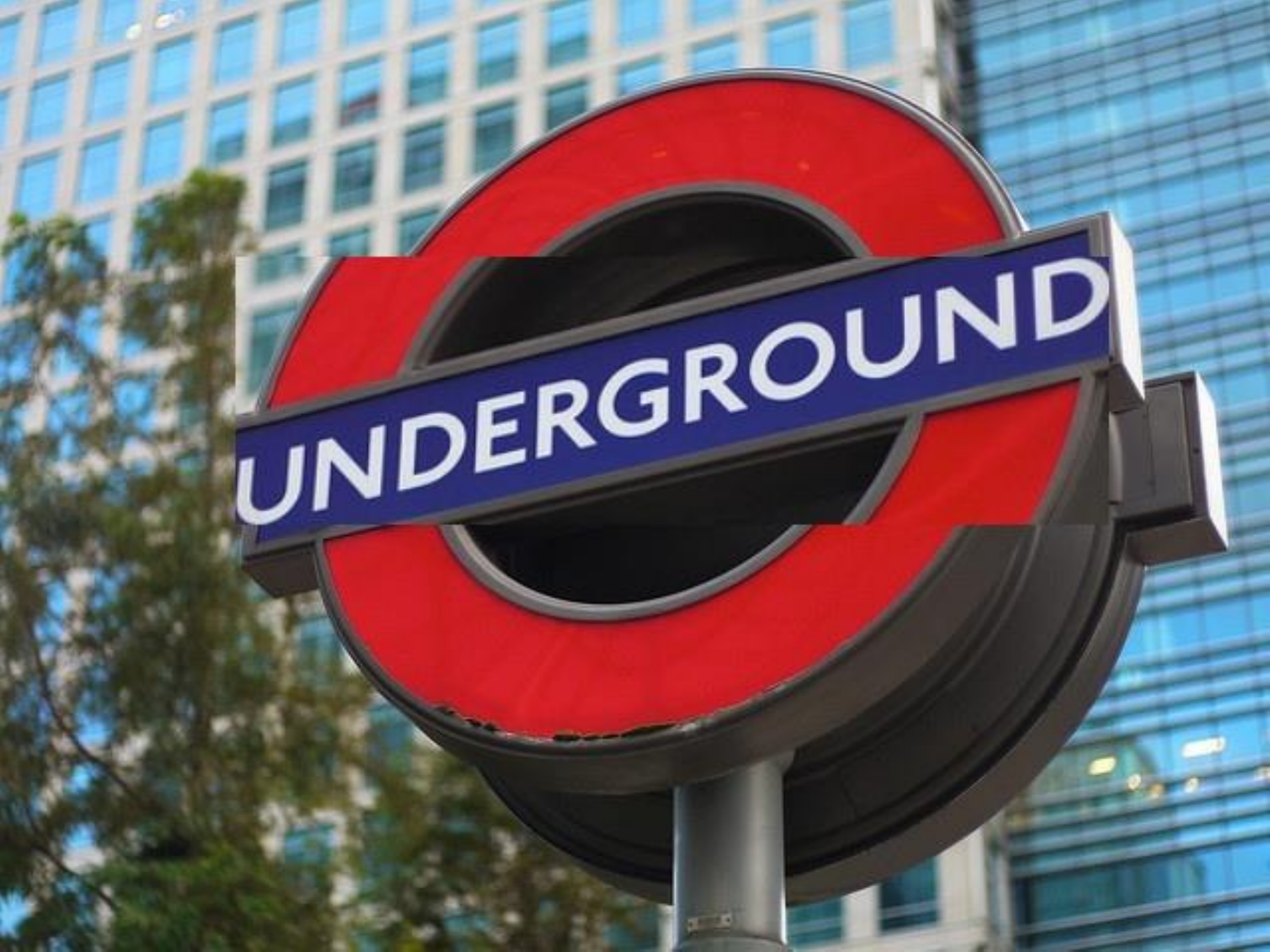}
}
\caption{Text detection region and its simple magnification in the natural scene image.}
\label{fig1}
\end{figure}

In this paper, we present a four-stage CNN-based text magnifier to expand the text in the image. The center position and content of text remains unchanged while the text are magnified. Simply magnifying the text detection regions results in background disturbance as shown in Fig.~\ref{fig1}. Instead, we propose a novel text magnifier to enlarge text in scene images without effect the background. It is composed of four task: character erasing, character extraction, character magnifying, and image synthesis. Character erasing aims to remove the text in the image while keep the background unchanged. This idea and framework is presented in the previous work~\cite{Nakamura2017Scene}, in which a Convolutional Neural Network (CNN)-based Encoder-Decoder is adopted in an end-to-end mode for erasing scene text. Then, the text erased image and the original image are input to the character extraction network and it returns only the character connected components of the original image and a binary character mask for indicating. Subsequently, they are further input to the character magnify network and the character magnified results and its mask image are outputted. Finally, the character magnified image combines with its corresponding mask image and the character erased image are input to the image synthesis stage to generate the final text magnified scene text image.

Since we only magnify the character in pixel level, the affection of background region can be ignored. For experiment, we set two magnifying rate, namely magnifications of 1.2 times and 1.5 times, to observe the influence on the increase of this parameter. The four stage networks are trained independently and are further fine-tuned end to end. The text magnified results are evaluated quantitatively by measuring the Structural Similarity (SSIM)~\cite{Zhou2004Image} between text magnified image and the processed ground truth.

The main contributions of our proposed method are presented as follows.

\begin{itemize}

\item First, to the best of our knowledge, we are the first to propose text magnifier application in the natural scene images without text recognition. The concept of enlarging the text region is useful since it can prevent missed reading and misunderstanding for special groups of people.

\item Second, the scene text magnifier is composed of four stages and can be fine-tuned end to end. It involves with multi-task networks and the intermediate process can provide text erasing and text extraction results, both of which are important research orientation in  CV field.

\item Finally, our proposed method focuses on processing character components instead of the text regions. The style and texture of text keeps unchanged. It can magnify the scene text effectively while not effect the background.

\end{itemize}

%The rest of the paper is structured as follows: A selection of related work is reviewed in Sect.~\ref{Sect2}. Sect.~\ref{Sect3} presents our proposed method in detail and the dataset we collected in introduced in Sect.~\ref{Sect4}. In Sect.~\ref{Sect5}, we give the experimental results which include the details of databases and the experiment implements. Finally, Sect.~\ref{Sect6} gives a summarization and conclusion of this paper.

\section{Related work}\label{Sect2}

\subsection{Scene text segmentation}

In recent years, a large number of deep learning-based scene text detection approaches have been proposed, most of which have been summarized in the latest survey~\cite{Long2018Scene}. Generally, these approaches can
be roughly divided into three groups: regional proposal-based~\cite{Ma2017Arbitrary}, anchor-based~\cite{Liao2018TextBoxes} and semantic segmentation-based~\cite{Dan2018PixelLink,Tang2017Scene}. The text extraction stage in our proposed method is most related to the semantic segmentation-based text detection, which aims to assign the pixel-wise text and non-text labels to an image.

%Yao et al.~\cite{} cast the detection task as a semantic segmentation problem, by predicting three kinds of score maps: text/non-text, character classes, and character linking orientations. Zhang et al.~\cite{} proposed to predict the saliency map by FCN to generate the TextBlocks, and character candidates are extracted using MSER. Tang et al.~\cite{} segmented the scene text by cascaded CNNs to extract both the edge and text components.

\subsection{Scene text erasing}

Scene text erasing is first presented in the work~\cite{Nakamura2017Scene}. The goal is to erase the text regions and make them hard to be detected. We used an inpainting deep neural network (DNN) converting the problem as image transformation refereing to transforming images from a source image space to a target image space. The inpainting DNN is considered as the eraser. It composes of Convolutional neural networks (CNN) in front and deconvolutional neural networks (DeCNN) subsequently to recover the image resolution. Previous methods for removing graphic texts are often dealt with  born-digital images~\cite{Modha2014Image,Wagh2015Text}. Their ability to remove scene texts are limited because scene texts undergo many distortions, such as uneven illuminations and perspective distortions. Most recently, the EnsNet~\cite{zhang2019EnsNet}, which was built on an end-to-end trainable FCN-ResNet-18 network with a conditional generative adversarial network (cGAN), was proposed. The feature of the former is first enhanced by a novel lateral connection structure and then refined by four carefully designed losses: multi-scale regression loss and content loss, which capture the global discrepancy of different level features; texture loss and total variation loss, which primarily target filling the text region and preserving the reality of the background. The latter is a novel local-sensitive GAN, which attentively assesses the local consistency of the text erased regions.

\subsection{Image synthesis}

Image synthesis refers to insert synthetic objects into existing photographs. Expertise process~\cite{Karsch2011Rendering} go through the geometry estimation, 3D scene computation including the consideration of the physical light, surface materials, and camera parameters. After fix the position of the scene, the objects are rendered and composted into the original image. For scene text image synthesis, the location to put the text and make it appear naturally is the most important. The SynthText dataset~\cite{Gupta2016Synthetic} is synthesized from acquiring text and natural scene images for text detection. Text magnifier requires to extract the text firstly, then magnify them and embed the processed text to the original image.

%It first segments the natural scene image into contiguous regions based on local colour and texture cues, and a dense pixel-wise depth map is obtained using the CNN~\cite{Liu2015Deep}. Then, for each contiguous region a local surface normal is estimated. Next, a colour for text and, optionally, for its outline is chosen based on the region��s colour. Finally, a text sample is rendered using a randomly selected font and transformed according to the local surface orientation; the text is blended into the scene using Poisson image editing~\cite{Rez2003Poisson}.

\section{The Proposed Method}\label{Sect3}

\begin{figure}[t]
	\begin{minipage}{\hsize}
		\begin{center}
			\includegraphics[width=9.0cm]{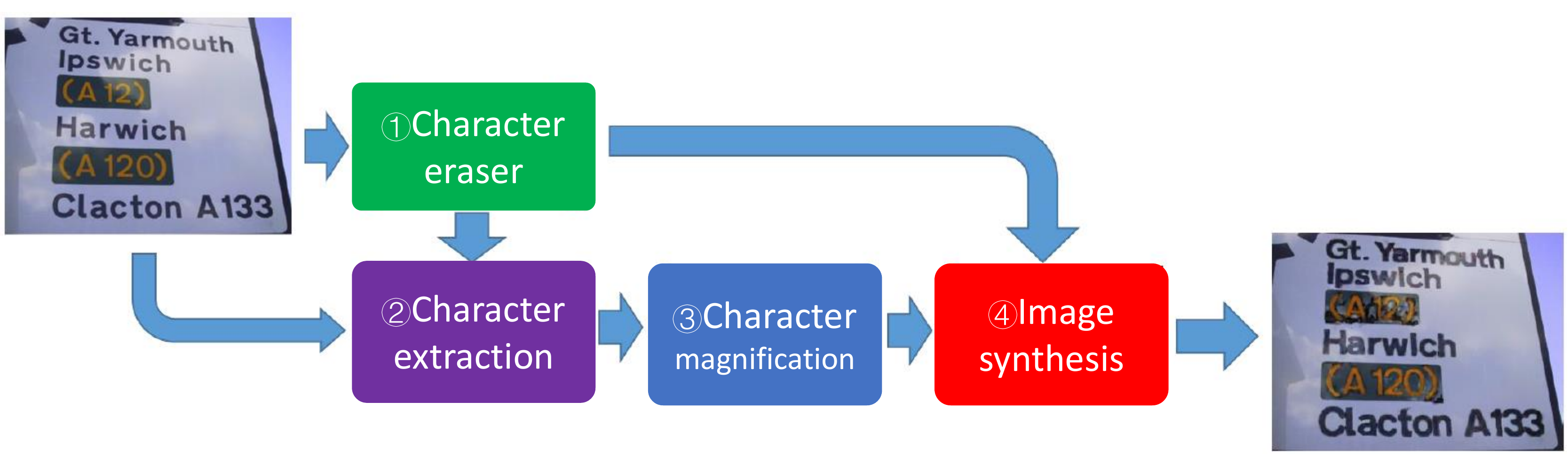}
			\caption{The structure of the proposed scene text magnifier.}
			\label{fig2}
		\end{center}
	\end{minipage}
\end{figure}

In this section, we introduce the four cascaded CNNs for text magnifier in detail. The framework is displayed in Fig.~\ref{fig2}, which is composed of four stages: character erasing, character extraction, character magnifying, and image synthesis. The network architecture of each stage is correspondingly displayed in Fig.~\ref{fig3}. Those networks are trained separately initially and fine-tuned in end-to-end manner consequently.

\subsubsection{First stage: character erasing network}
The character erasing network uses the hourglass encoder-decoder CNNs with 4 layers of convolution and deconvolution symmetrically. The details of the architecture and data for training can be found in the previous work~\cite{Nakamura2017Scene}. It inputs the original scene text image and outputs an image with text removed and background being unchanged as exampled in Fig.~\ref{fig3}(a). The output of text erased image is then reused as an input image for character extraction at the second stage and also for image synthesis at the fourth step.

\subsubsection{Second stage: character extraction network}
Character extraction network targets to extract the text components from the original image. Since the process in first stage removes the text and keep the background unchange, it can assist to extract the text by inputting both the original image and the character erased image. Therefore, we design the network as shown in Fig.~\ref{fig3}(b). Convolutional features are extracted from the original and character erased images respectively, and then are concatenated. Two branch of Deconvolutional process outputs the images with character components and a character mask. The character components keeps the color of the text in original image, and the background is unified in black. The mask image is a binary image which indicates the character in pixel level. The results are used as the input image of character magnify in the third stage.

\subsubsection{Third stage: character magnify network}
Character magnify network (as shown in Fig.~\ref{fig3}(c)) involves with two input images, namely the character components and mask, and features are extracted through two branch of convolution layers. After integrating the two features, character magnifying is performed by deconvolution process and the magnified text and its mask which means the same as mentioned in character extraction network are outputted. The magnified text image and its mask is synthesized with the erased image by through CNN-based image synthesis at the fourth step.

For text magnifying, several variations on the position of the magnified characters are conceivable. For example, magnifying text while keeping the interval between characters will not produce characters overlap. But the initials and tail of the text may become invisible and result in misunderstand of the meaning. Oppositely, magnifying text while maintaining the center position for each character may lead to characters overlap if the interval between characters is narrow. However, since it is possible to fill all the magnified characters in the image, it becomes easier to understand the word even after magnifying. Therefore, we select the second strategy to magnify characters in the image without changing the original center position of each character.

\subsubsection{Fourth stage: image synthesis network}
\begin{figure}[t]
\centering
\subfloat[Character erasing network.]
{\label{fig3_1}
\includegraphics[width=8.2cm,height=2.2cm]{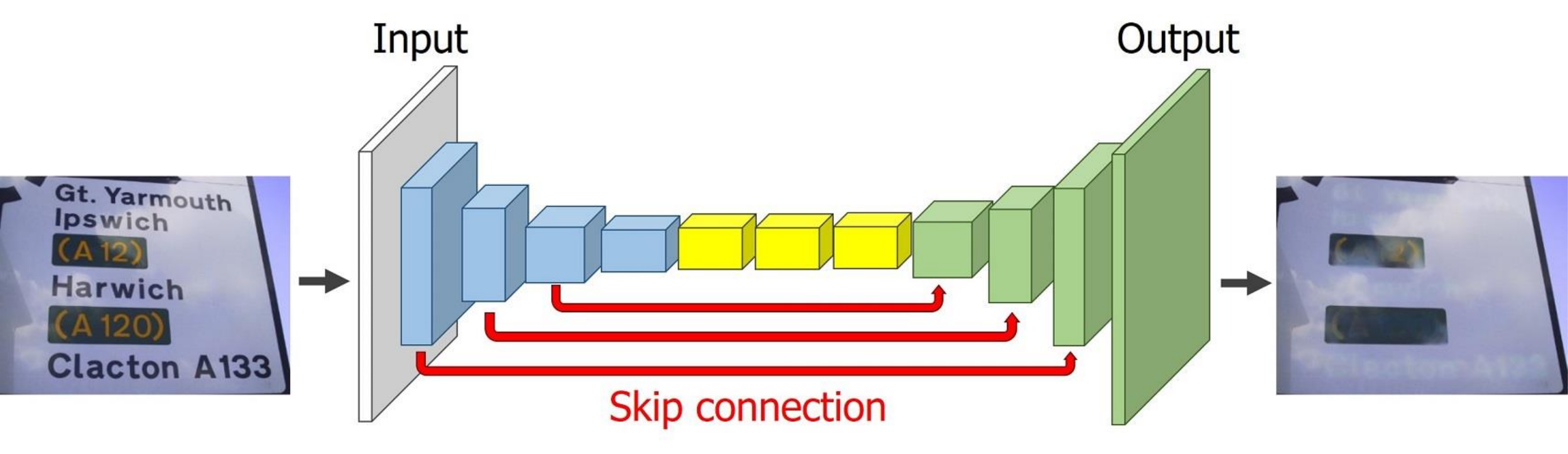}
}\\
\subfloat[Character extraction network.]
{\label{fig3_2}
\includegraphics[width=8.0cm,height=3.0cm]{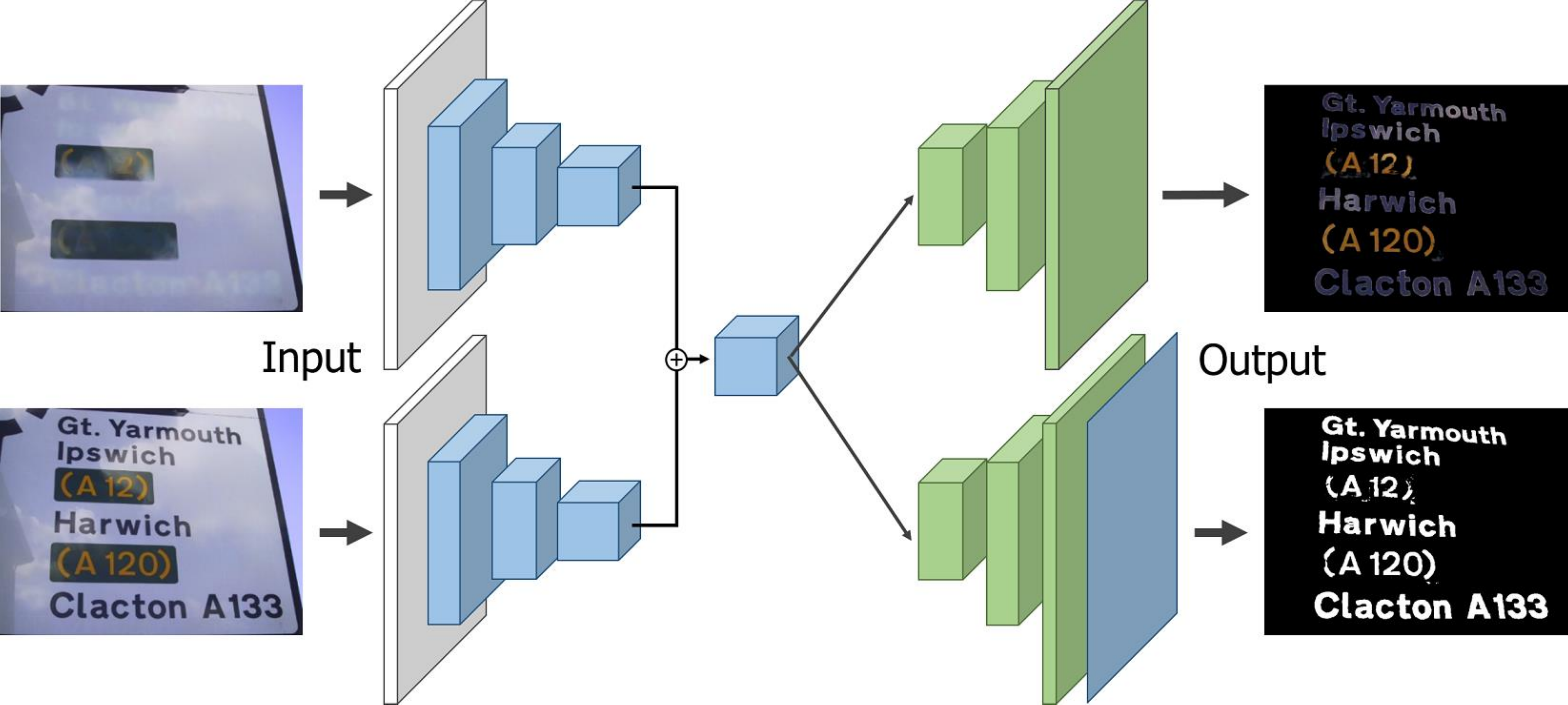}
}\\
\subfloat[Character magnifying network.]
{\label{fig3_3}
\includegraphics[width=8.5cm,height=3.3cm]{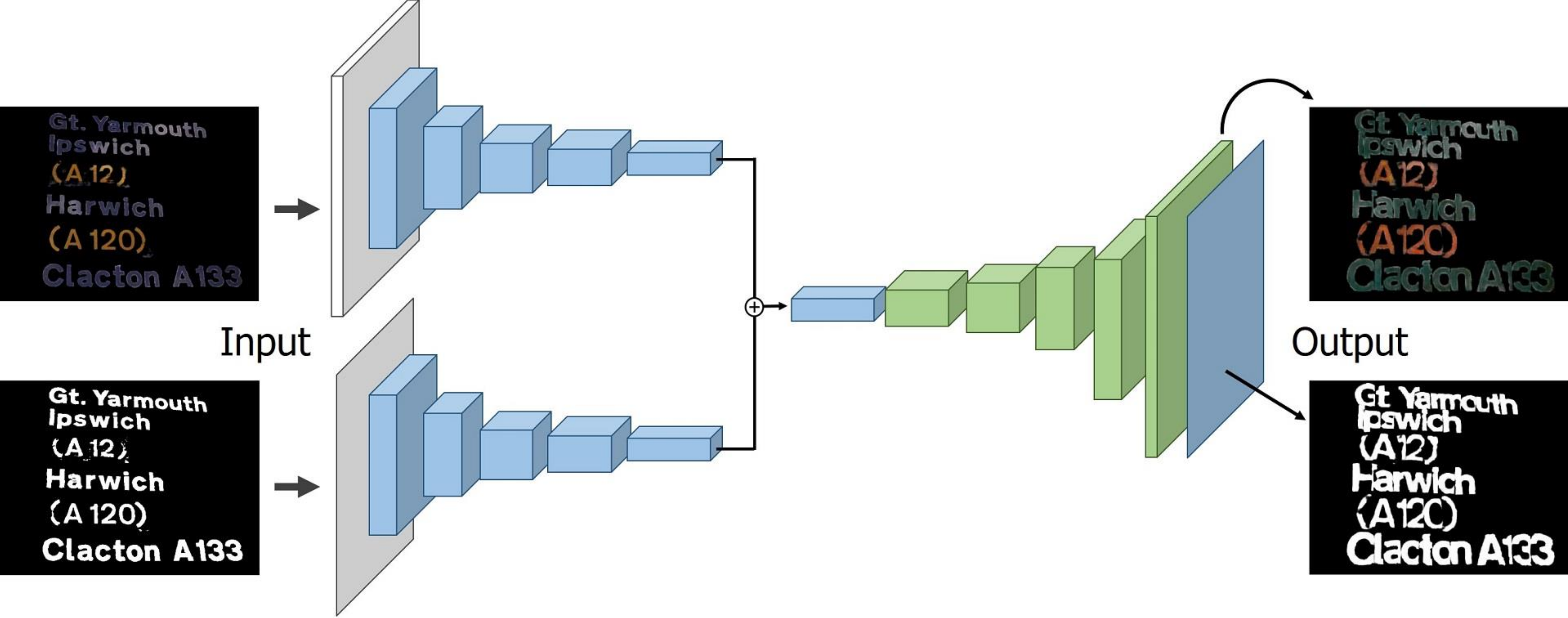}
}\\
\subfloat[Image synthesis network.]
{\label{fig3_4}
\includegraphics[width=7.5cm,height=4.7cm]{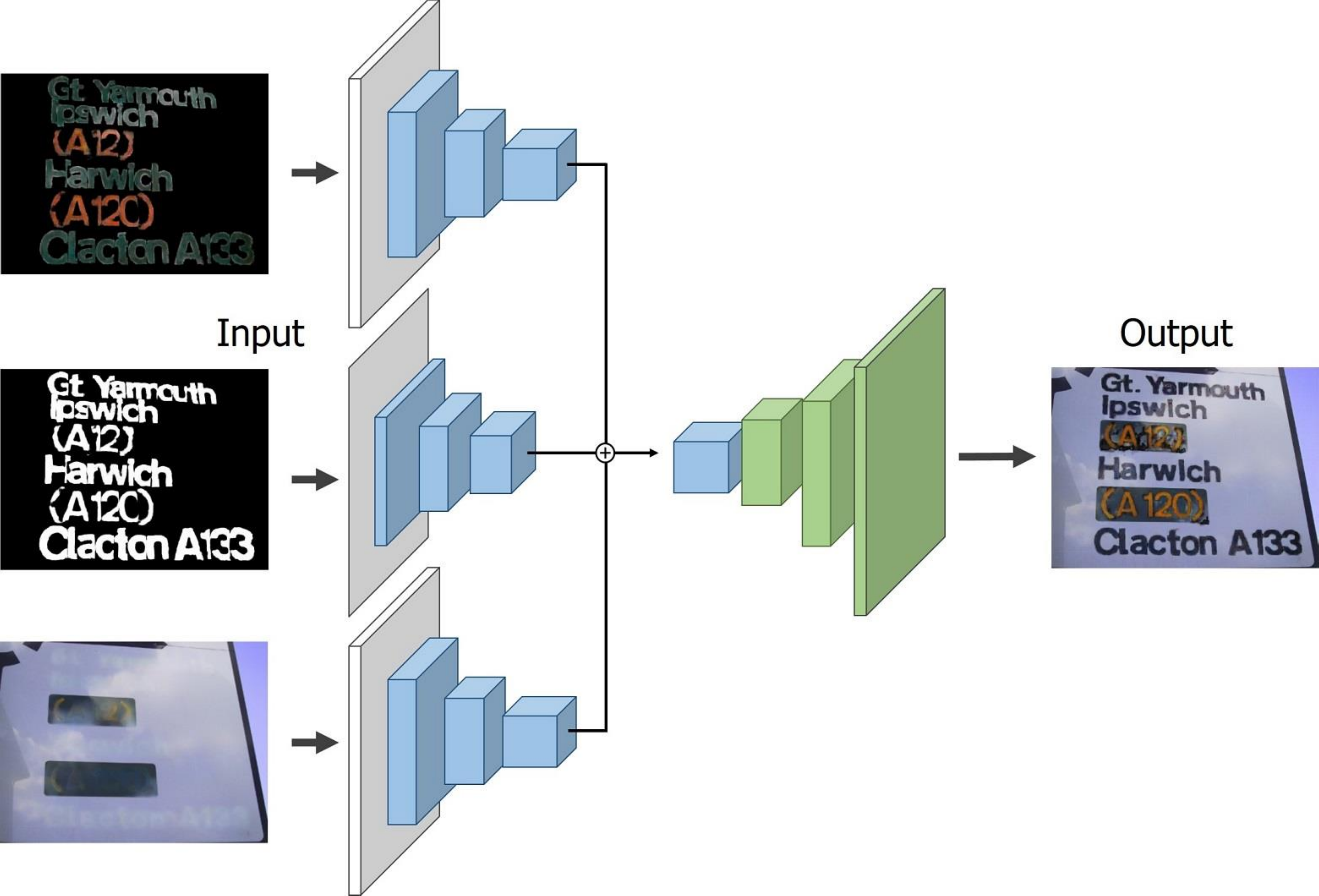}
}
\caption{The structure of each stage in text magnifier.}
\label{fig3}
\end{figure}

The image synthesis network, as shown in Fig.~\ref{fig3}(d), inputs the magnified character image, its corresponding mask and text erased image. The features are separately extracted through three CNNs, and then integrated and decoded to the final result with magnified characters attached on the natural scene image.

\section{Data for Training}\label{Sect4}

In the training phase, the four networks are trained independently and then fine-tuned in an end-to-end manner. These 4 CNNs use Batch Normalization and ReLU as the activation function as in the case of a simple Encoder-Decoder, and the activation function in the last layer of the three former CNNs is Sigmoid. For image synthesis network, the mean square error is used as the loss function.

\begin{figure}[t]
	\begin{minipage}{\hsize}
		\begin{center}
			\includegraphics[width=9.0cm]{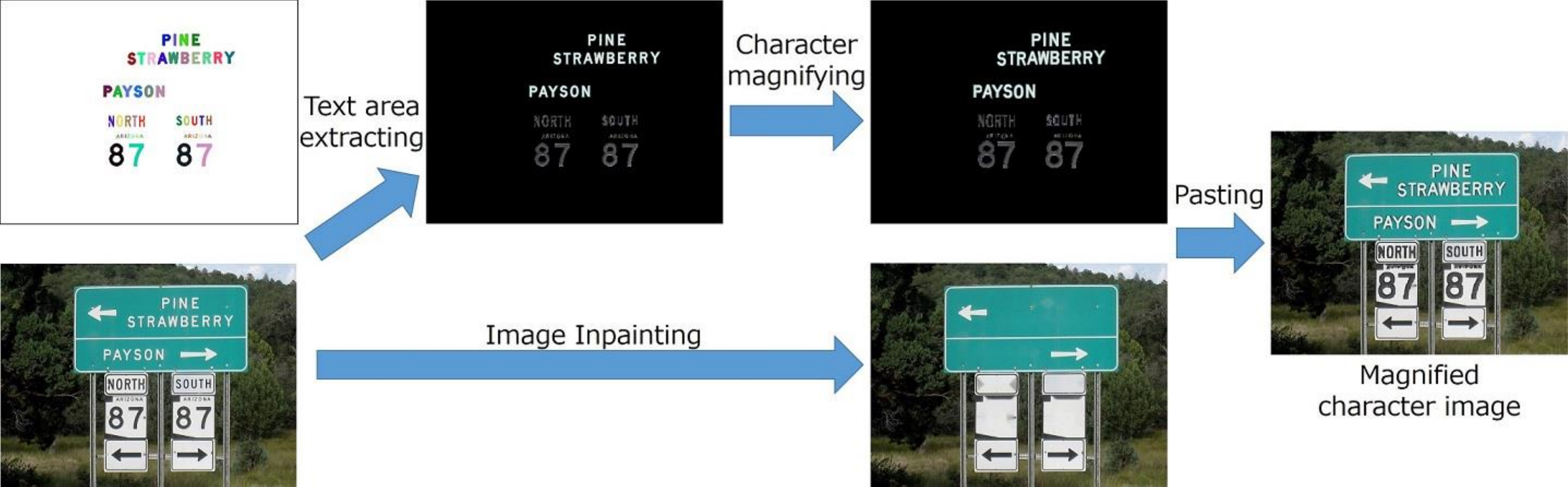}
			\caption{The procedure of training data generation.}
			\label{data}
		\end{center}
	\end{minipage}
\end{figure}

Images in ICDAR2013 and Flickr datasets are used for training data generation. Text in these two datasets are mostly focused and attached on signboards, billboards, etc., with different orientation. In our method, four types of data are required: scene text erased images, character segmentation, magnified character segmentation and character magnified images. Among them, the character segmentation ground truth in contained in the two public datasets. The generation of the data follows the process flow in Fig.~\ref{data}.

First, the original image and the annotation of character image are used to extract the character components. On the other hand, the text erased images are produced by inpainting process. Next, the extracted character component is magnified by enlarge the character bounding box region with the designated magnification rate. Since we do not change the center position of each character region, the magnified character components will be relocated in its corresponding original place of the text erased image. Two relocation of the magnified characters are discussed in the experiment section. In total, we get 3247 images for training and 233 images for test.

In addition, in order to confirm the influence on the increase of the magnifying rate, two types of magnification rate of 1.2 times and 1.5 times are used. The number of learning iterations in each stage was set to 1000. After that, the 300 iteration of fine-tuning is performed.

\section{Experimental Results}\label{Sect5}

\subsection{Qualitative evaluation}

\begin{figure}[t]
	\begin{minipage}{\hsize}
		\begin{center}
			\subfloat[\label{a}]{
				\includegraphics[width=8.5cm]{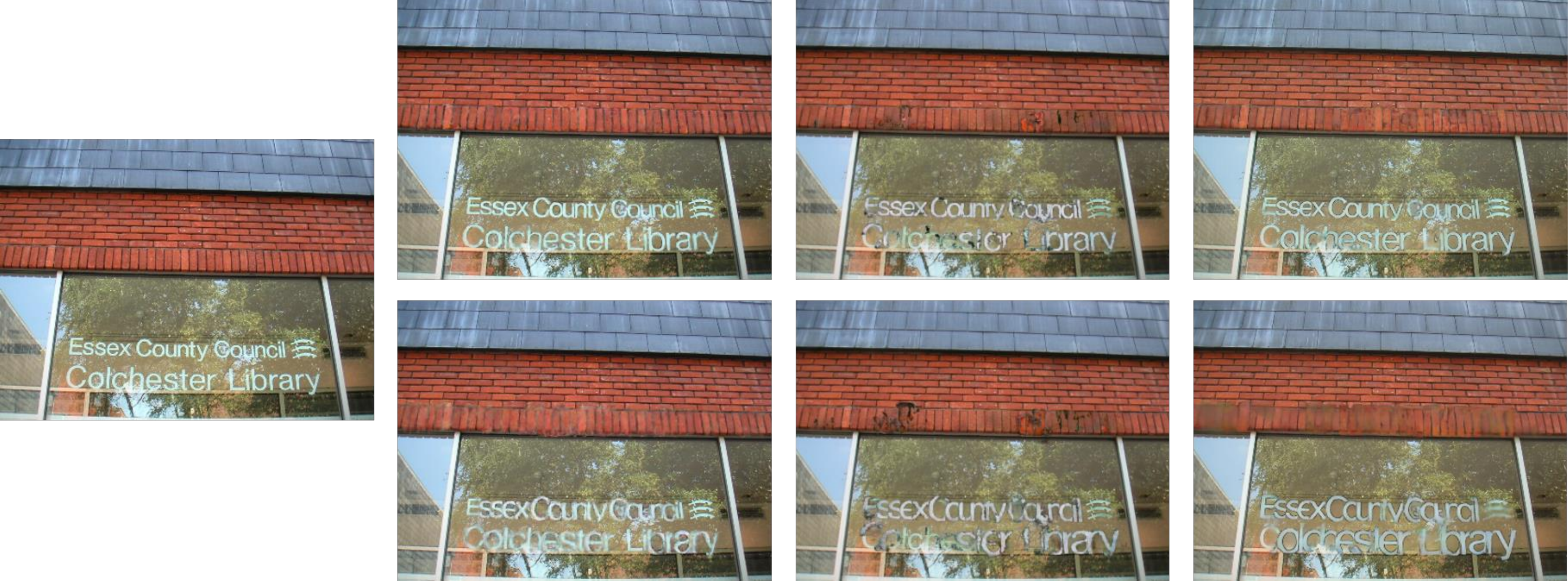}
			}
			\\
			\subfloat[\label{b}]{
				\includegraphics[width=8.5cm]{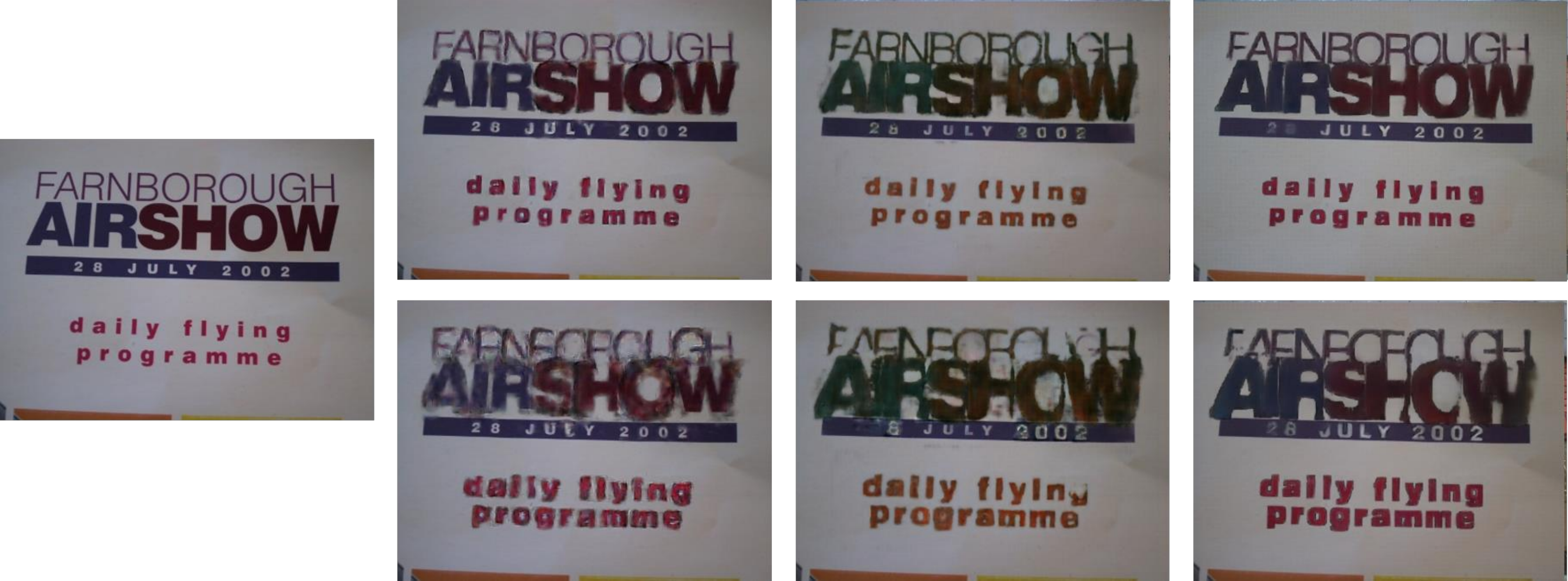}
			}
            \\
            \subfloat[\label{c}]{
				\includegraphics[width=8.5cm]{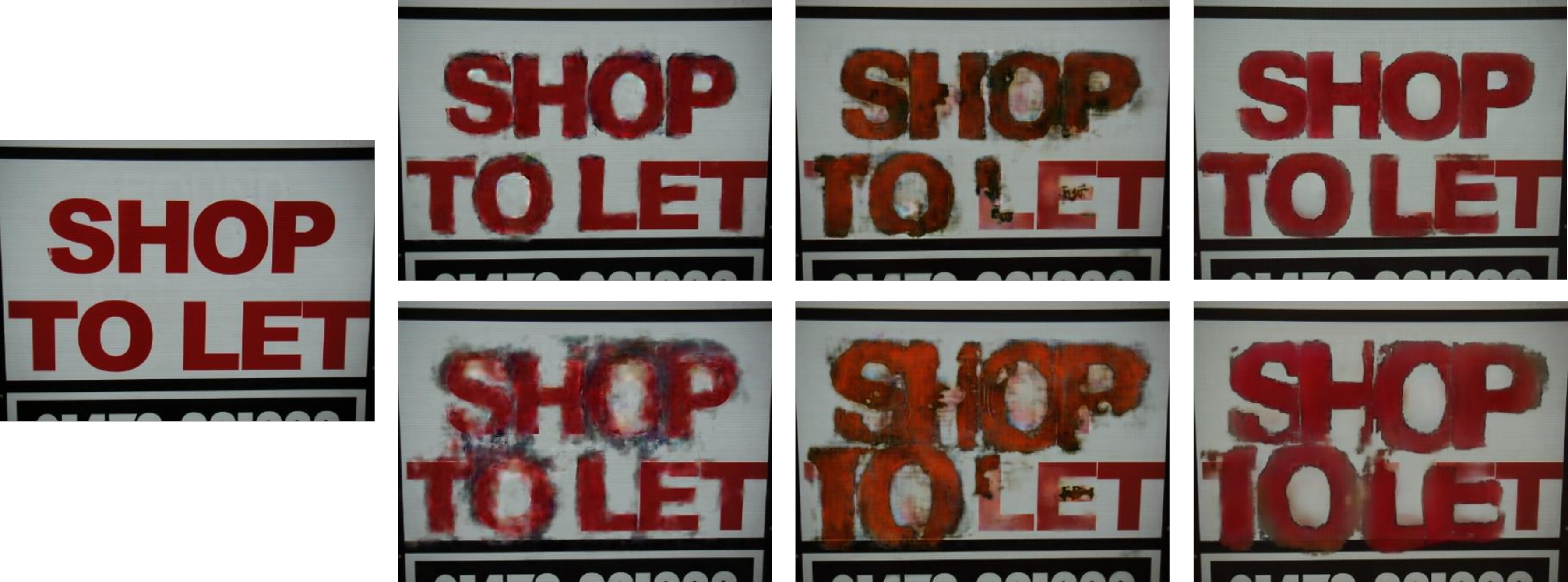}
			}
			\caption{Example of character magnifying results. The leftmost for each example presents the original image. The images in the upper row are the results magnified by 1.2 times and the lower row magnified by 1.5 times. The single Encoder-Decoder results, four-stage magnifying results and fine-tuned results are displayed from left to right.}
			\label{fig4}
		\end{center}
	\end{minipage}
\end{figure}

Some results of experiment are shown in Fig.~\ref{fig4}. The single hourglass encoder-decoder for text magnify is also performed and the results are dropped here for comparison with our proposed four-stage text magnifier. From the results in Fig.~\ref{fig4}(a), we can find that the shape of the magnified character is the most clear by using the 4-stage text magnifier plus fine-tune. The results without fine-tune may result in black color appearing around the character. The result from the encoder-decoder network shows that the shape of the character is incomplete and some characters are not responded during magnifying.

In the image of Fig.~\ref{fig4}(b), the word of ``AIRSHOW'' is blurred by using 1.5 times magnification rate through the encoder-decoder network. Oppositely, the outline of the character is very clear by our method and the improvement is obvious specially after fine tune. By observing the results in Fig.~\ref{fig4}(c), we can find that our method can magnify small-size characters better than large-size characters. Since less overlap of neighbored characters, the clearer text region is obtained. Additionally, our method has better ability to magnify the large-size character compared to encoder-decoder network.

\subsection{Quantitative evaluation}

To evaluate the performance quantitatively, we measure the SSIM of magnified result in each character bounding box region with the ground truth. Except the mentioned three method above, SSIM was also measured on the image outputted by simply performing character detection, magnifying the region and pasting it on the original image. Since the pixel-level ground truth of character detection regions in ICDAR2013 dataset has released, the magnifying process is performed on those detected regions directly. They are further pasted to the original position by aligning with the center axis of the labeled character components bounding boxes.

The ground truth to measure the SSIM here is the image obtained by magnifying only the character segmentation parts, and the background remains to be unchanged. The averaged SSIM value measured by simple encoder-decoder network, four-stage networks, fine-tune result and the detection-based mehtod are shown in Table \ref{resSSIM}.
Among the three CNN-based methods, the SSIM of the 4-stage network plus fine-tune gets the highest at the magnification rate of both 1.2 times and 1.5 times. In combination of the qualitative analysis, which demonstrates the simple encoder-decoder network is unstable and may cause blurring for magnifying text and the incompliant results of black holes may appear in the image by using only four-stage network without fine-tune, this comparison result further confirms quantitatively that the shape of the character after magnifying by our proposed method is more clearer and can keep more original textness features compared to the other two methods.

\begin{figure}
	\begin{center}
		\subfloat[\label{SSIMexpa}]{
			\includegraphics[width=8.5cm,height=2.8cm]{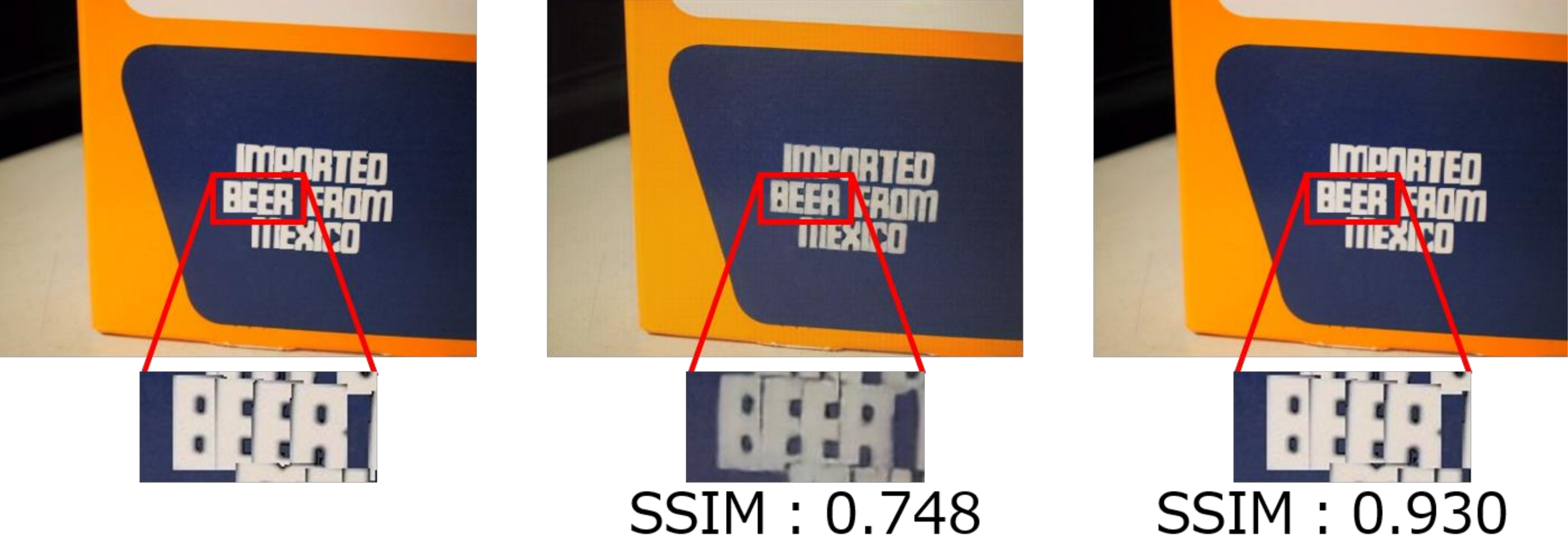}
		}
		\\
		\subfloat[\label{SSIMexpb}]{
			\includegraphics[width=8.5cm,height=2.8cm]{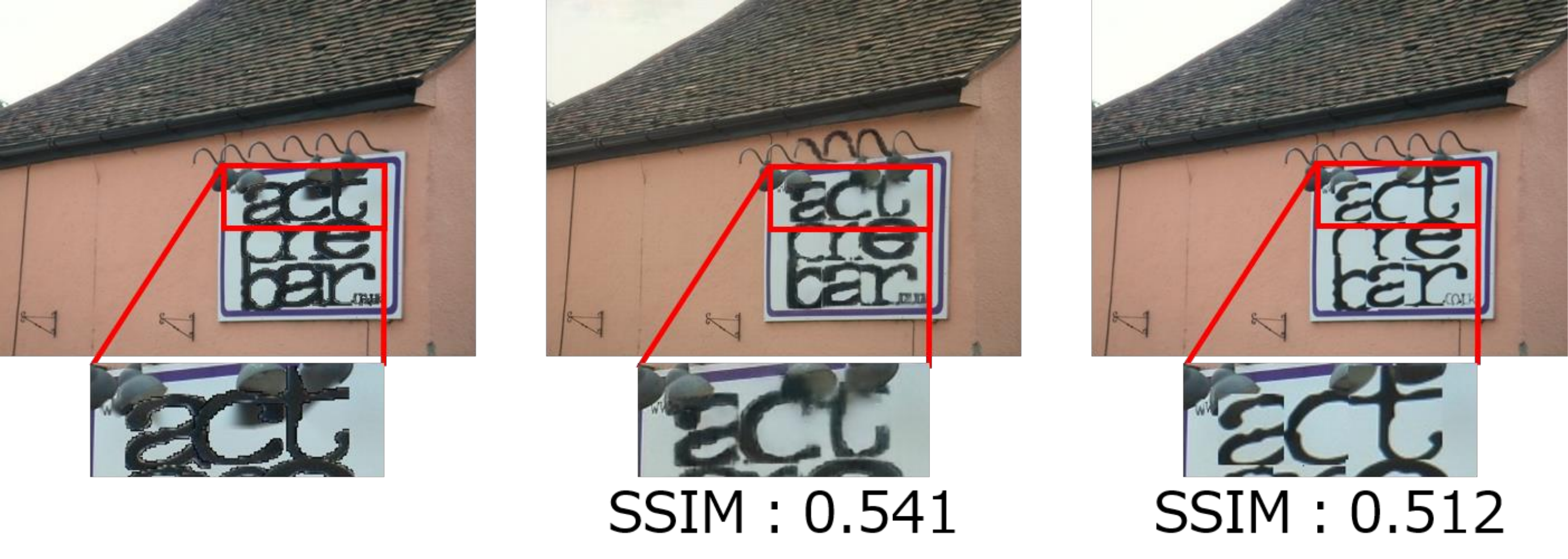}
		}
		\caption{Examples of magnified results and SSIM socres in some text regions.
		The most left images show the ground truth and the compared text regions. Images in the mid and right show the the results by 4-stage text magnifier network plus fine-tuning and the result of character detection/ pasting, respectively. The compared region is highlighted and its SSIM score is measured.}
		\label{fig5}
	\end{center}
\end{figure}

In the character detection-based magnification method, the SSIM score is higher than our proposed method.
It rises from the reason that most of the text regions in ICDAR2013 dataset are attached on signboard or advertisement as shown in Fig.~\ref{fig5}(a). The background color is simple. Most characters are focused with large size. No cluster around the characters. After magnifying and pasting, the characters do not occlude other objects in the background region. So, the SSIM can get high score.
On the other hand, when there is an object around the character like ``act'' of Fig.~\ref{fig5}(b), the detection-based magnification method hides the object because it magnify each background around the character. After pasting, the magnified character regions will occlude background. However, the character part can be magnified without hiding the object by our method, and the SSIM is higher than the detection-based magnification method.

\begin{table}
	\centering
	\caption{SSIM measurement score in character bounding box regions of magnified text image.}
	\label{resSSIM}
	\begin{tabular}{|c|c|c|}
		\hline
		Character magnifying method         & magnifying rate & SSIM\\ \hline
		\multirow{2}{*}{Simple Encoder-Decoder type CNN}	& 1.2 & 0.574 \\
		& 1.5 & 0.470 \\ \hline
		\multirow{2}{*}{4-steps CNN without Fine-tuning}				& 1.2 & 0.544 \\
		& 1.5 & 0.482 \\ \hline
		\multirow{2}{*}{4-steps CNN with Fine-tuning}		& 1.2 & 0.617 \\
		& 1.5 & 0.550 \\ \hline
		\multirow{2}{*}{Detection-based magnification}			& 1.2 & 0.729 \\
		& 1.5 & 0.704 \\ \hline
	\end{tabular}
\end{table}

\subsection{Discussion of training data}

\begin{figure}[!t]
	\begin{minipage}{\hsize}
		\begin{center}
			\subfloat[]{
				\includegraphics[clip,width=0.46\hsize]{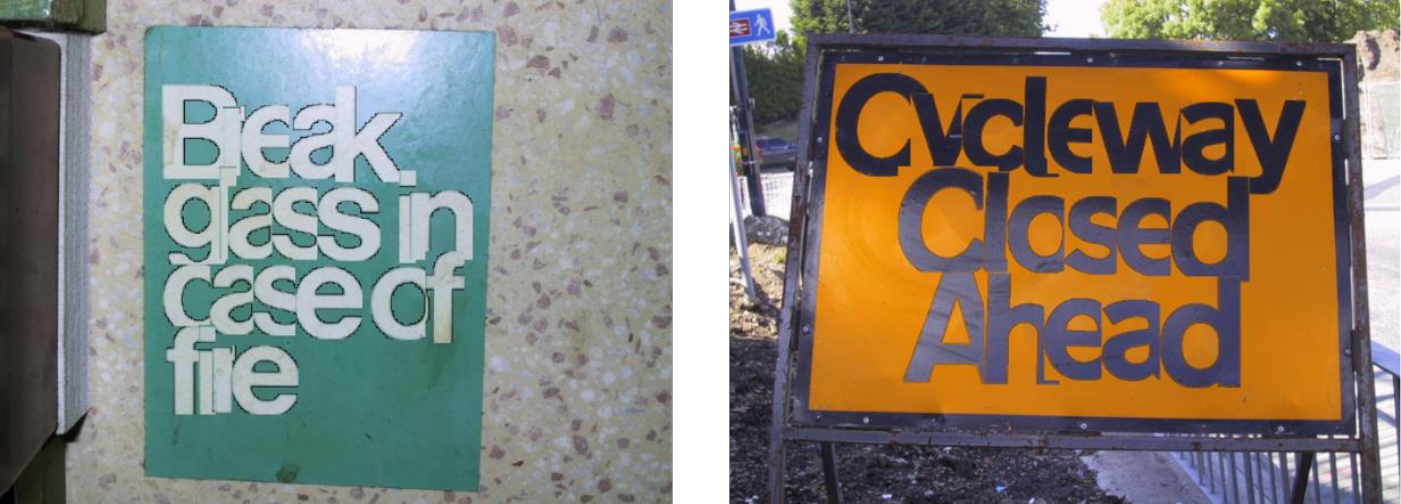}
			}
			\subfloat[]{
				\includegraphics[clip,width=0.46\hsize]{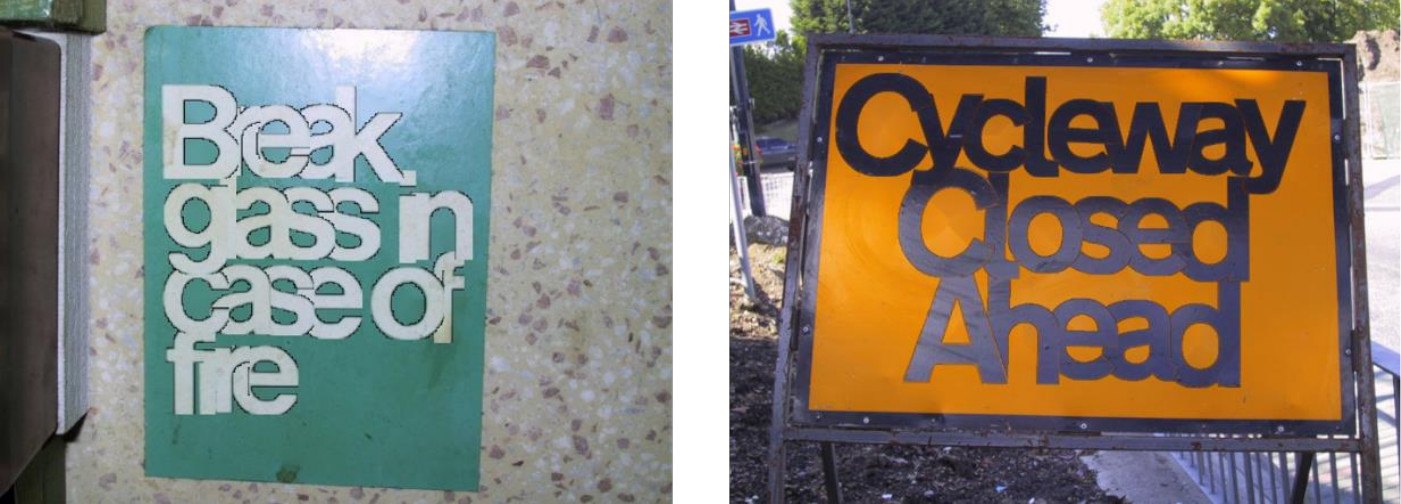}
			}
			\caption{Examples of text magnified training samples.
				(a) The character is pasted in the rectangular area with priority given to the lower right character.
				(b) The character is pasted in the character component parts with the priority given to the upper left character.}
			\label{magnumber}
		\end{center}
	\end{minipage}
\end{figure}

There are two ways to generate the text magnified training data. One method is to enlarge the character bounding box regions and paste the enlarged regions to image in left-to-right order as shown in Fig.~\ref{magnumber}(a). It brings overlapping problem, especially for characters have narrow intervals. Since the magnified region is rectangular, the background part surrounding the magnified character is also pasted to the image and will conceal the real background. This situation also happens on using magnified word-level text detection regions. The other way to generate the training data is based on pixel-level character annotation. And if the magnified characters are overlapped during pasting, we give the priority to the character at the upper left as shown in Fig.~\ref{magnumber}(b). With this process, the background around the magnified character will not be pasted, so each character will not be chipped from the overlap.

%As a result, there is a possibility that the character in the upper left is overwritten with the background, and part of the character is missing. The readability drops by training with this kind of data.

\begin{figure}[t]
	\begin{minipage}{\hsize}
		\begin{center}
			\subfloat[\label{resdataseta}]{
				\includegraphics[width=8.0cm, height=1.7cm]{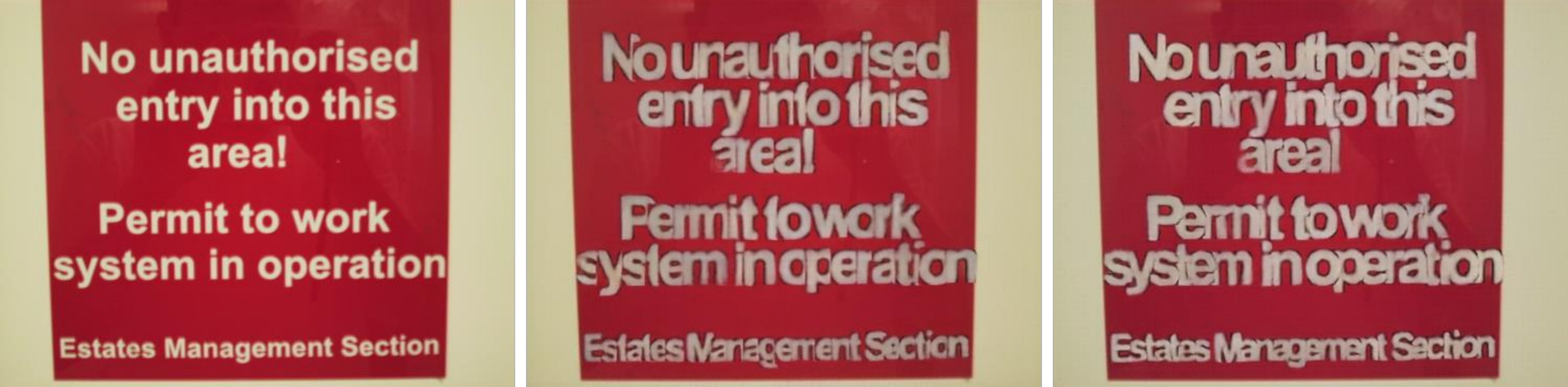}
			}
			\\
			\subfloat[\label{resdatasetd}]{
				\includegraphics[width=8.0cm, height=1.7cm]{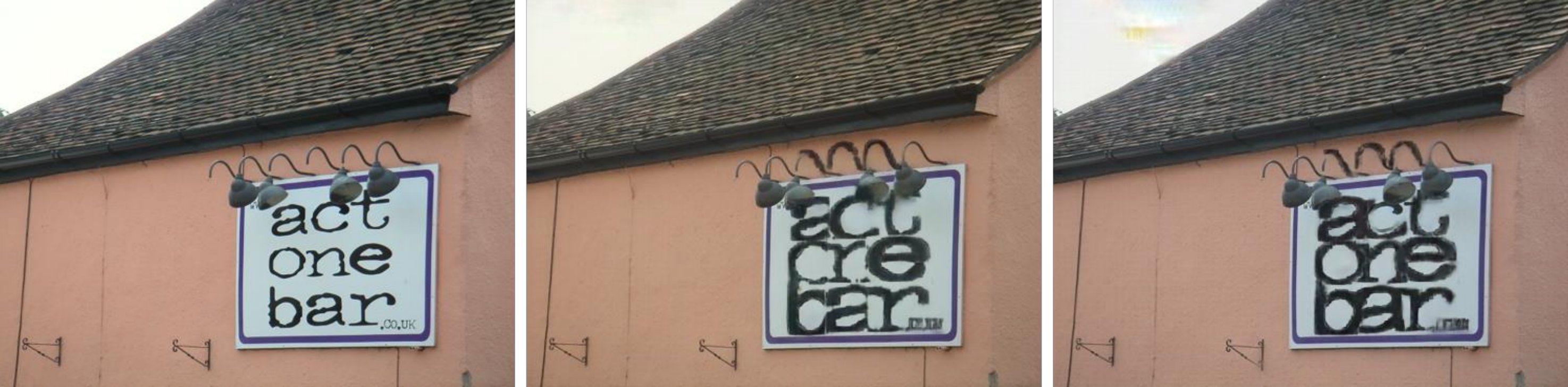}
			}
			\\
            \subfloat[\label{resdatasetb}]{
				\includegraphics[width=8.0cm, height=1.7cm]{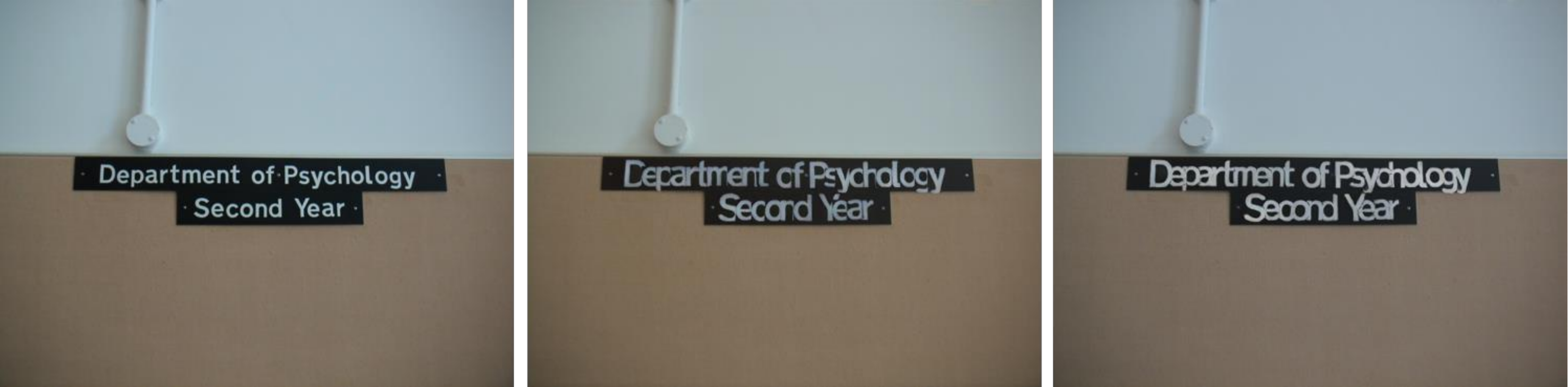}
			}
			\caption{Example of the character magnified results by different training data.
			The columns from the left to right display the original images, the results with first kind of training data, and the results with second kind of training data, respectively.}
			\label{numberres}
		\end{center}
	\end{minipage}
\end{figure}

Some comparison results are shown in Fig.~\ref{numberres}. From left to right, the images are the original images, the magnified results by using the first category of data and the results by using the second category of data. Both results are obtained with 1.5 times magnification rate by our proposed method.

In Fig.~\ref{numberres} (a) and (b), the magnified character components of the word ``into'' and
``one'' are overlapped due to the narrow blank space between characters. If the first kind of training data are used, the character of `t' in the word ``into'' appears to be `f' and the word ``one'' becomes ``cne''. As a result they may be incorrectly recognized.
When we train the network by the second kind of data, characters will not be occluded by background. From the result, we can see that the magnified character on the left has higher priority and covers part of its neighbored right character if they have overlap.

The character occluding gets more serious in the results of Fig.~\ref{numberres} (c) when we use the first kind of training data. Most right part of characters are missing or occluded by its right magnified character bounding box. By using the second kind of training data, it can avoid this situation. The black background does not move onto the character, and each of the character is magnified while keeping its original shape. The readability of the characters after magnifying will increase.

%In this way, by changing the order of the arrangement of the characters so as not to include the background in the rectangular area of each character during magnifying, it is understood that the character can be magnified with keeping its readability.

\subsection{Discussion of character magnification strategy}

In the above proposed method, the character magnification is performed without changing the centroid of each character. That may result in character overlap. If the characters are mainly on the center of images and we magnify them in word-level while keeping the interval between characters, the results may be more clearer and readable. Therefore, we observe the results of magnifying only the character part with the center of the original image as a reference.

%In our proposed method, we modify the third stage of character magnification strategy. First, a direct character magnification method without using CNN is applied. The output of the second stage, namely the character components image, is extended to larger size based on the magnification rate. Then, only the the centered area with the same size of the original image is cropped from the magnified character components image. After that, it is input to the image synthesis stage for final result.

\begin{figure}[t]
\centering
\subfloat[CoordConv.]
{\label{coordconv}
\includegraphics[width=8.0cm]{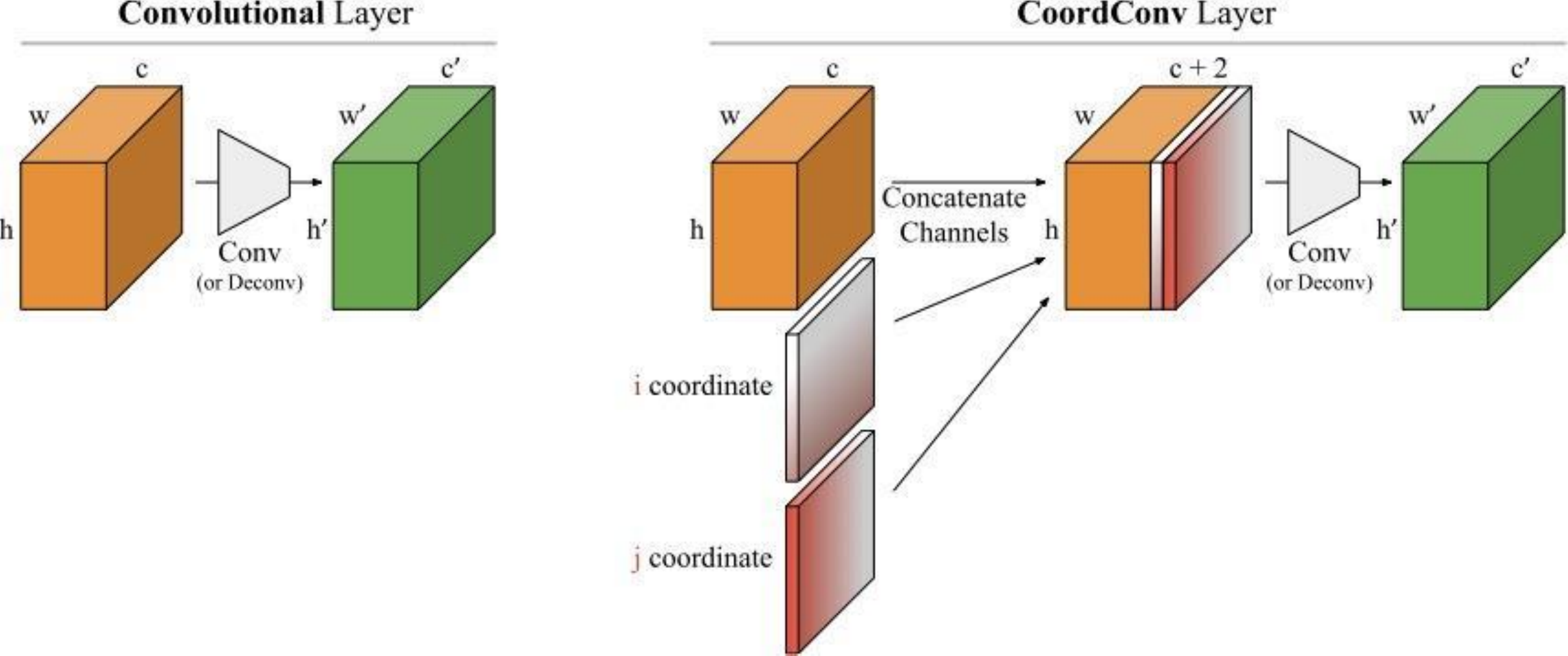}
}\\
\subfloat[The CoordConv-based character magnify network.]
{\label{net3}
\includegraphics[width=8.0cm,height=2.8cm]{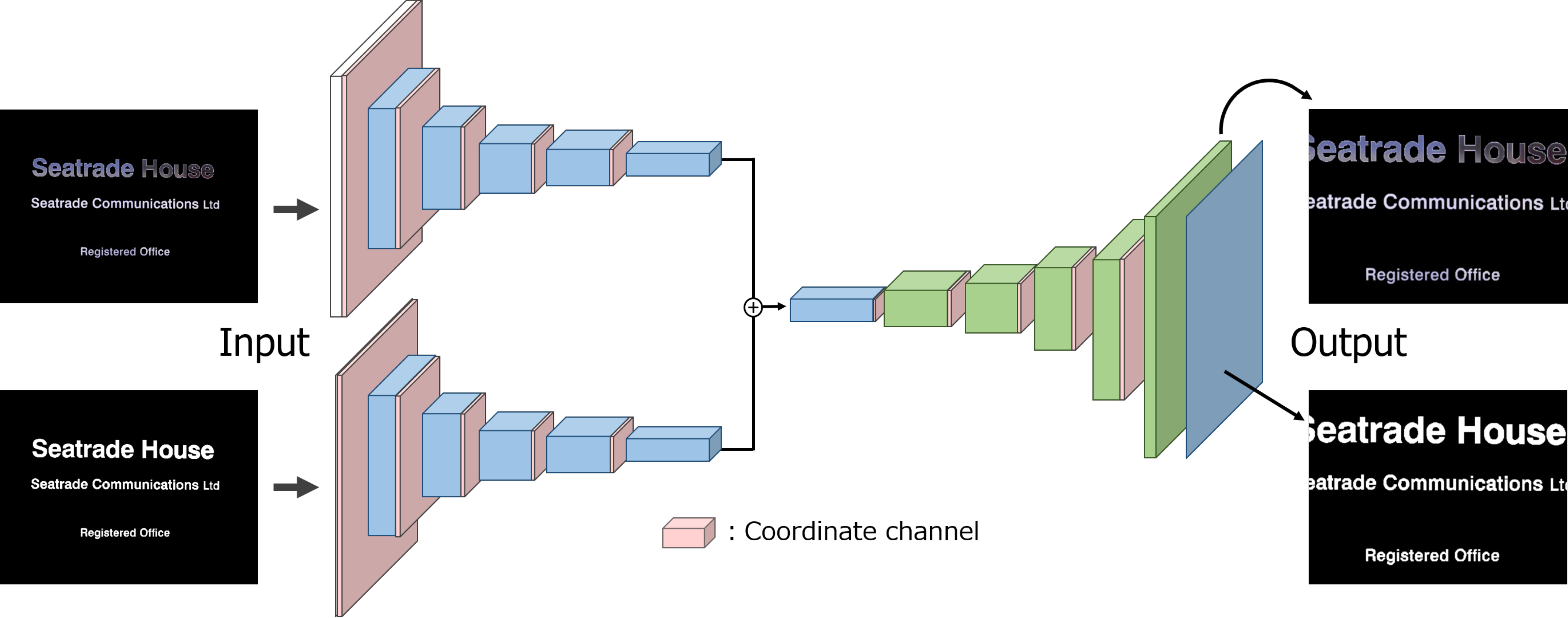}
}
\caption{The replaced network for character magnification in the third stage by using CoordConv layers.}
\label{figconv}
\end{figure}

In order to train the whole model in end-to-end way, we replace the network in the third stage by CoordConv-based~\cite{Liu2018An} CNN (Fig.~\ref{figconv}(b)) for image center-based character magnification. The CoordConv adds the channels indicting the x and y coordinates of the image in convolutional layers as shown in Fig.~\ref{figconv}(a). Since magnifying character based on the center of the image involves with the character movement, using the CoordConv by giving the coordinates in the image makes each convolution layer learn the shifted distance of each pixel on x and y-axis direction. Therefore, the shifted location of magnified characters can be determined by computing the coordinate information of CoordConv.

\begin{figure}
\centering
\subfloat[]
{\label{final_1}
\includegraphics[width=2.0cm]{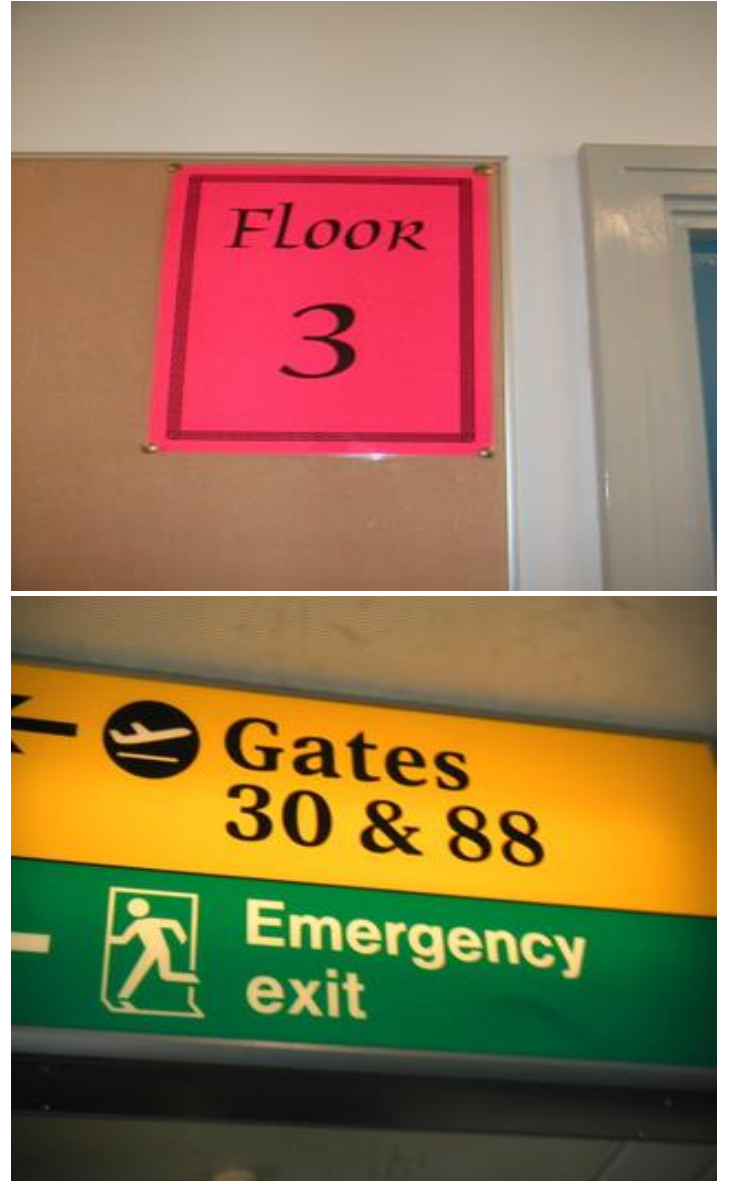}
}
\subfloat[]
{\label{final_2}
\includegraphics[width=2.0cm]{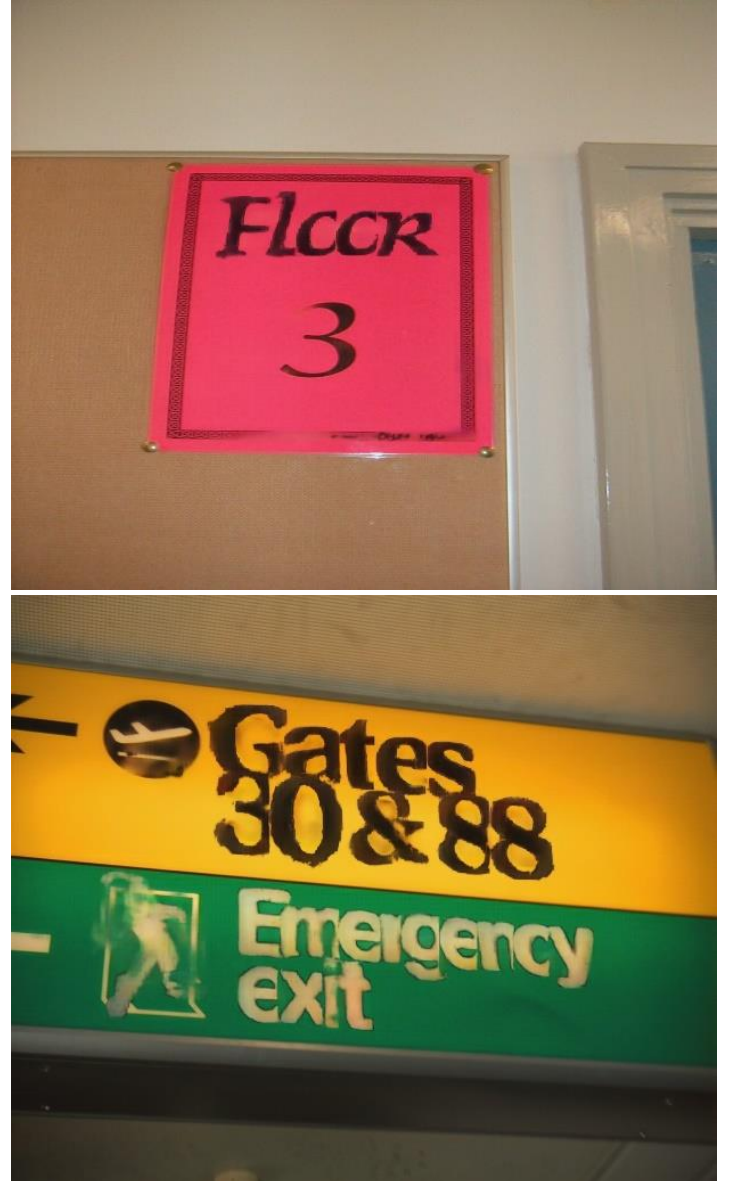}
}
\subfloat[]
{\label{final_3}
\includegraphics[width=2.0cm]{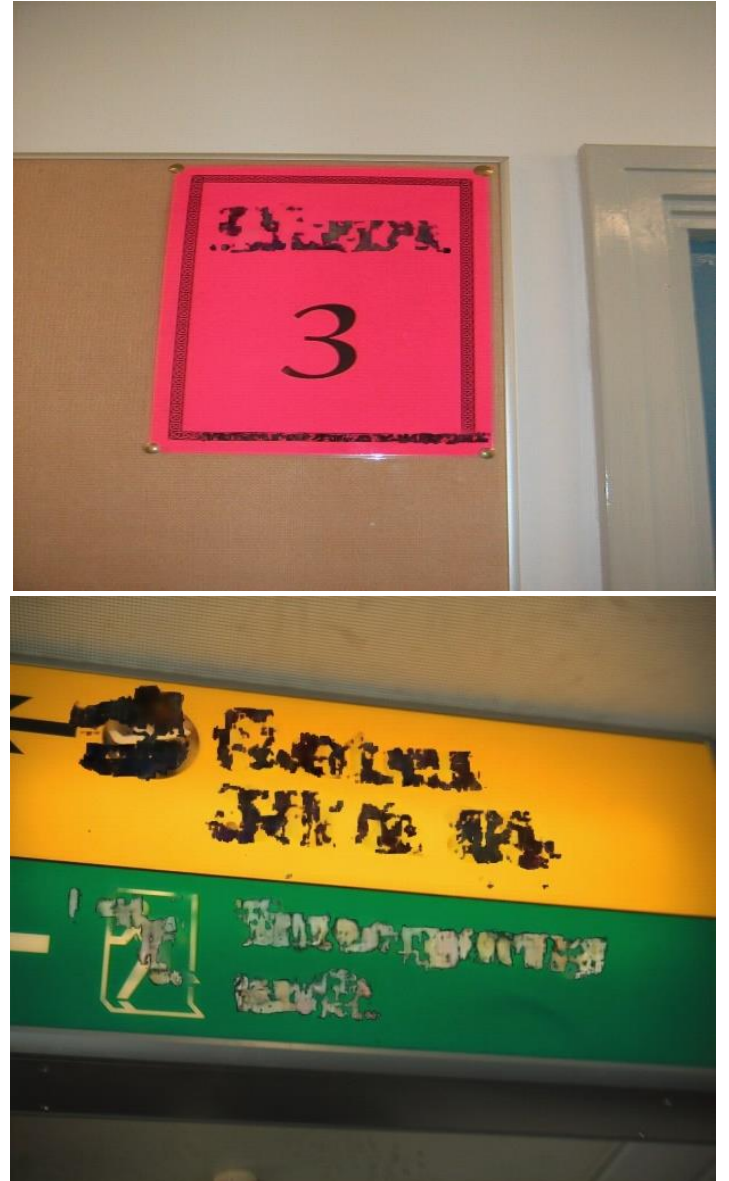}
}
\subfloat[]
{\label{final_4}
\includegraphics[width=2.0cm]{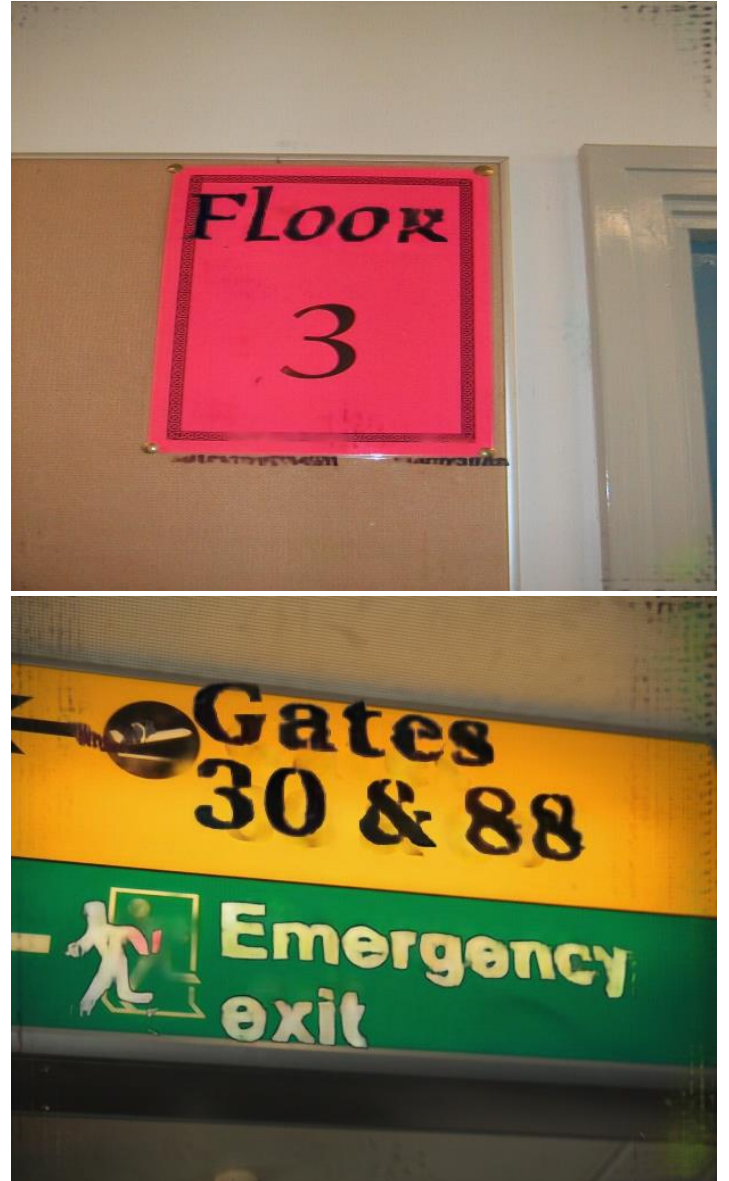}
}
\caption{The text magnify results. (a) The original images. (b) The magnified results based on character bounding box center. (c) The magnified results based on image center without Coordconv.(d) The magnified results based on image center by Coordconv-based CNN.}
\label{fig_final}
\end{figure}

Results of character magnification by different methods are shown in Fig.~\ref{fig_final}. As analyzed in the above subsection, the magnified characters may overlap if they are synthesized based on original character bounding box center. Fig.~\ref{fig_final}(c) are the results of only using CNN without Coordconv to magnify characters based on image center. We can see it fails for this task. This is because characters are moved when they are magnified. If there is no position information to tell where each pixel of the characters should move to within the range of the filter in each convolution layer, it is difficult to magnify the characters to a proper location and maintain their original shapes. The Coordconv-based magnification method can solve the above problems as shown in Fig.~\ref{fig_final}(d). It works well if the characters are mainly located in the center of the image. Since it magnifies characters based on image center, if the characters are distributed at the image border, some characters may disappear after image synthesis.

\section{Conclusion}\label{Sect6}
In this paper, we designed a scene text magnifier, which aimed to magnify the text in natural scene images for assisting people who had myopia or dyslexia. It was composed of four sub-networks integrating scene text erasing, text extraction, text magnify and image synthesis. They were all built on the convolution and symmetric deconvolution neural networks. After independent training for each network, they were cascaded and fine-tuned in end-to-end manner. The SSIM measurement was performed to evaluate the magnified results quantitatively and explained the disadvantage of just magnifying text detection results which changed the background and brought the occlusion. The ways to train and magnify the characters were discussed. Finally, we came to conclude that our proposed method can magnify scene text effectively without effecting the background by magnifying the pixel-level character annotation based on its original center location.

\section*{Acknowledgments}

This work was partly supported by the JSPS KAKENHI Grant Number JP17K19402, JP17H06100 and National Natural Science Foundation of China under Grant 61703316.

\bibliographystyle{IEEEtran}
\bibliography{anna}

\end{document}